\documentclass{bmvc2k}

\title{\texttt{DAVINCI}: A Single-Stage Architecture for Constrained CAD Sketch Inference}

\addauthor{Ahmet Serdar Karadeniz}{ahmet.karadeniz@uni.lu}{1}
\addauthor{Dimitrios Mallis}{dimitrios.mallis@uni.lu}{1}
\addauthor{Nesryne Mejri}{nesryne.mejri@uni.lu}{1}
\addauthor{Kseniya Cherenkova}{kcherenkova@artec-group.com}{1,2}
\addauthor{Anis Kacem}{anis.kacem@uni.lu}{1}
\addauthor{Djamila Aouada}{djamila.aouada@uni.lu}{1}

\addinstitution{
Interdisciplinary Centre for Security, Reliability and Trust (SnT), \\
University of Luxembourg
}
\addinstitution{
Artec 3D, \\
Luxembourg
}

\runninghead{Karadeniz et al.}{DAVINCI}

\def\eg{\emph{e.g}\bmvaOneDot}

\usepackage[utf8]{inputenc} %
\usepackage[T1]{fontenc}    %
\usepackage{hyperref}       %
\usepackage{url}            %
\usepackage{booktabs}       %
\usepackage{amsfonts}       %
\usepackage{nicefrac}       %
\usepackage{microtype}      %
\usepackage{xcolor}         %
\usepackage{todonotes}
\usepackage{lipsum}
\usepackage{todonotes}
\usepackage{stmaryrd}
\usepackage{multirow}
\usepackage{amsmath}
\usepackage{subcaption}
\usepackage{wrapfig}
\usepackage{amssymb}
\usepackage{enumitem}

\begin{document}

\maketitle

\begin{abstract}

This work presents \texttt{DAVINCI}, a unified architecture for single-stage Computer-Aided Design (CAD) sketch parameterization and constraint inference directly from raster sketch images.
By jointly learning both outputs, \texttt{DAVINCI} minimizes error accumulation and enhances the performance of constrained CAD sketch inference. Notably, \texttt{DAVINCI} achieves state-of-the-art results on the large-scale SketchGraphs dataset~\cite{seff2020sketchgraphs}, demonstrating effectiveness on both precise and hand-drawn raster CAD sketches. To reduce \texttt{DAVINCI}'s reliance on large-scale annotated datasets, we explore the efficacy of CAD sketch augmentations. We introduce \textit{Constraint-Preserving Transformations} (CPTs), \emph{i.e.} random permutations of the parametric primitives of a CAD sketch that preserve its constraints. This data augmentation strategy allows \texttt{DAVINCI} to achieve reasonable performance when trained with only $0.1\%$ of the SketchGraphs dataset. Furthermore,
this work contributes a new version of SketchGraphs, augmented with CPTs. The newly introduced \textit{CPTSketchGraphs}\footnote{\url{https://github.com/cvi2snt/CPTSketchGraphs}}  dataset includes 80 million CPT-augmented sketches, thus providing a rich resource for future research in the CAD sketch domain.
\end{abstract}

\begin{figure*}[h]
    \setlength{\belowcaptionskip}{-0.5cm}
    \centering
    \includegraphics[width=\textwidth]{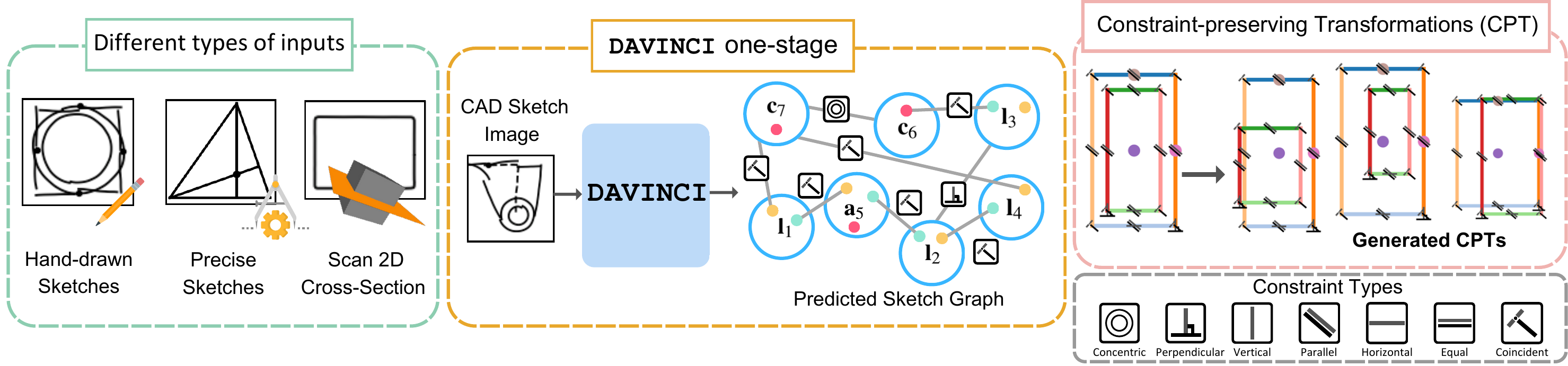}
    \vspace{-0.1cm}
    \caption{We propose \texttt{DAVINCI}, a novel single-stage network for constrained CAD sketch parameterization. \texttt{DAVINCI} effectively parameterizes different types of input sketches, from precise to hand-drawn as well as 2D cross-sections. We also introduce Constraint-Preserving Transformations (CPTs), \emph{i.e.} an augmentation strategy tailored to constrained CAD sketches.}
    \label{fig:teaser}
\end{figure*}

\section{Introduction}

The field of industrial manufacturing has experienced a significant transformation over the past few decades, a change largely driven by the adoption of Computer-Aided Design (CAD)~\cite{Zhang2009DesignII,Otey2018RevisitingTD,seff2022vitruvion,seff2020sketchgraphs}. At the centre of modern CAD workflows are CAD sketches, typically consisting of a collection of 2D parametric primitives (\eg lines, arcs, circles, etc.). The CAD designing process is usually initiated by drawing these CAD sketches that can be subsequently expanded to form 3D solids via CAD operations, such as extrusions or revolutions. Commonly, CAD designers will also define relationships between sketch primitives in the form of constraints, dictating how the parametric primitives of the sketch adjust in response to modifications (\eg perpendicular constraint between two lines)~\cite{para2021sketchgen,seff2020sketchgraphs,seff2022vitruvion}. CAD research highlights the importance of effective constraint practices on sketch development~\cite{Kyratzi2020ACF, Company2019OnTR, Kyratzi2022IntegratedDI}, as constraints are a widely agreed upon mechanism to represent design intent~\cite{Zhang2009DesignII, Otey2018RevisitingTD}. 

The recent availability of large-scale datasets of CAD sketches such as \textit{SketchGraphs}~\cite{seff2020sketchgraphs} and \textit{CAD as a Language}~\cite{ganin2021computer} has spurred a line of research into various CAD sketch-related applications. These include novel sketch synthesis~\cite{ganin2021computer,para2021sketchgen,seff2022vitruvion}, auto-completion~\cite{seff2022vitruvion}, auto-constraining~\cite{seff2020sketchgraphs, para2021sketchgen,seff2022vitruvion}, concept discovery~\cite{yang2022discovering}, and constrained CAD sketch inference from raster sketch images~\cite{para2021sketchgen, seff2022vitruvion}. The focus of this work is on the latter challenge: \textit{constrained CAD sketch inference from raster sketch images}. Given the critical role of sketches in the CAD industry, the ability to infer parametric constrained CAD sketches from raster sketch images has a significant potential to streamline labor-intensive processes. This \textit{reverse engineering} capability is applicable across a diverse range of inputs that are often adopted within CAD workflows. As shown in the left panel of Figure~\ref{fig:teaser}, these range from precise technical drawings to informal hand-drawn sketches that encapsulate the initial conceptualization of a design, as well as 2D cross-sections~\cite{cross_sections} — two-dimensional slices typically obtained in reverse engineering software by taking the intersection of a plane with a 3D scan.

This work proposes \texttt{DAVINCI}, a novel transformer-based architecture for simultaneous CAD sketch parameterization and constraint inference from sketch raster images. Departing from existing methods~\cite{para2021sketchgen, seff2022vitruvion}, which address primitive and constraint inference as distinct problems, \texttt{DAVINCI} employs a single-stage architecture (see mid panel of Figure~\ref{fig:teaser}). By predicting both primitives and constraints within a unified network, it effectively reduces the error accumulation associated with multi-stage processes, thereby achieving state-of-the-art performance on the large-scale SketchGraphs dataset~\cite{seff2020sketchgraphs}. In order to reduce the reliance of transformer-based architectures (including \texttt{DAVINCI}) on large-scale annotated dataset availability, this work explores data augmentation techniques tailored to CAD sketch applications. While data augmentation is commonplace in vision pipelines, they are less straightforward and often overlooked for CAD-related tasks. The proposed CAD sketch augmentation technique leverages the constraints of input sketches to create new CAD sketches preserving the constraints but with different parameterizations. This is achieved by importing constrained CAD sketches into CAD software (\eg FreeCAD~\cite{FreeCAD}) and automatically applying  \textit{Constraint-Preserving Transformations} (CPTs) on them to generate different parameterized CAD sketches. As shown in the right panel of Figure~\ref{fig:teaser}, the resulting CAD sketch augmentations have the advantage of bringing diversity, yet preserving the original distribution of CAD sketches.

\vspace{0.1cm}
\noindent \textbf{Contributions:} The main contributions of this work can be summarized as follows: 
\vspace{-0.2cm}
\begin{enumerate}[itemsep=2pt, parsep=0pt]
    \item \texttt{DAVINCI} is a novel architecture for single-stage CAD sketch parameterization and constraint inference from CAD sketch raster images.
    \item The proposed \texttt{DAVINCI} achieves state-of-the-art CAD sketch parameterization performance on the SketchGraphs dataset~\cite{seff2020sketchgraphs}. We extensively evaluate our method on both precise and hand-drawn CAD sketch raster images. We also report preliminary qualitative results showcasing the applicability of \texttt{DAVINCI} to 2D cross-sections.
    \item We propose a novel CAD sketch augmentation technique that leverages Constraint-Preserving Transformations (CPTs) of existing CAD sketches to generate new CAD sketches with identical constraints but different parameterizations. By applying CPTs to the original SketchGraphs dataset, which consists of 1.5 million CAD sketches, this work introduces an augmented version, the \textit{CPTSketchGraphs} dataset, containing 80 million CAD sketches.

\end{enumerate}

\noindent
The rest of the paper is organized as follows; Section~\ref{sec:related_work} reviews the related works. The proposed \texttt{DAVINCI} is detailed in Section~\ref{sec:method}. Section~\ref{sec:cpt} explains the proposed CPT-based augmentation. An experimental validation of the proposed framework is provided in Section~\ref{sec:experiments}. In Section~\ref{sec:conclusion}, we provide a conclusion and future works.

\section{Related Work}
\label{sec:related_work}

\noindent
\textbf{CAD Sketches.}
\textit{Feature-based} CAD modelling is currently a predominant paradigm for CAD model design~\cite{mallis2023sharp}. Within this framework, CAD sketches act as foundational elements for constructing complex design \textit{features} (\eg extrusions, holes, slots, bosses, and fillets). Recent reverse engineering pipelines have focused on the recovery of CAD sketches from diverse inputs. A line of work~\cite{wu2021deepcad, multicad, Khan_2024_CVPR} explores the inference of CAD model as a sequence of sketch-extrude operations by analysing point-clouds. Authors in \cite{cherenkova2023sepicnet} extract the parametarized primitive edges of a CAD model also from the corresponding point-cloud and \cite{dupont2022cadops} recover 2D CAD sketches from faces of the Boundary-representation (B-Rep) -- \emph{i.e.} a CAD representation for the geometric boundary of a 3D object. Concurrently to this work, \cite{karadeniz2024picasso} is proposed for CAD sketch inference from raster images of precise and hand-drawn CAD sketches. Note that all these methods neglect the recovery of parametric constraints that define relationships between primitives (\eg coincident, parallel, tangent, etc.), a component that is instrumental for further editing of CAD models via CAD software~\cite{Solidworks, Onshape, FreeCAD}.

\vspace{0.1cm}
\noindent
\textbf{Constrained CAD Sketch Inference.} A few recent methods~\cite{ganin2021computer,para2021sketchgen,seff2022vitruvion} tackle image-conditioned sketch parameterization and constraint inference, generally by exploring separate two-stage pipelines. Authors in \cite{ganin2021computer} follow a generic language modeling approach to generate the protocol buffer representation of a CAD sketch. \cite{willis2021engineering} introduces an autoregressive transformer for the generation of parametric primitives, whereas \cite{para2021sketchgen,seff2022vitruvion} also enable constraint inference via separate autoregressive constraint prediction networks. Related to ours is also the feed-forward method of SketchConcepts~\cite{yang2022discovering}, which operates on parametric primitives for joint concept discovery and auto-constraining. Authors show that SketchConcepts can be further extended to condition concept generation on images via a Vision Transformer (ViT)~\cite{Dosovitskiy2020AnII} encoder. In contrast to all these works, \texttt{DAVINCI} is specifically designed to jointly infer primitives and constraints from different types of CAD sketch raster images.

\vspace{0.1cm}
\noindent
\textbf{Graph Inference from Images.} Constrained CAD sketches can be viewed as graphs $G=(\mathcal{P},\mathcal{C})$ where nodes $\mathcal{P}$ constitute a set of parametric primitives and edges $\mathcal{C}$ denote designer-imposed geometric relationships or constraints~\cite{seff2020sketchgraphs}. A recent work~\cite{seff2022vitruvion} has explored the development of separate generative models for $\mathcal{P}$ and $\mathcal{C}$. This model decomposition is common in vision pipelines, such as in top-down human pose estimation~\cite{xiao2018simple, sun2019deep, yang2021transpose} or 2D to 3D keypoint lifting~\cite{tekin2017learning, martinez2017simple, li2019generating}. However, there has been growing interest in addressing traditionally two-staged problems with single-stage architectures to reduce error propagation and enhance efficiency~\cite{tekin2016structured, sun2017compositional, zanfir2018deep}. Particularly for the problem of Visual Relationship Detection (VRD)~\cite{dhingra2021bgt, lu2021context, chen2022reltransformer}, recent studies~\cite{cong2023reltr, shit2022relationformer} explore one-stage transformers to obtain scence graphs by jointly detecting objects and their relationships from images. Our work aligns with this research direction, as we propose the first single-stage transformer for joint CAD sketch parameterization and constraint prediction, represented as a graph, from CAD sketch raster images.

\vspace{0.1cm}
\noindent
\textbf{CAD Sketch Augmentations.}
A few augmentation strategies have been proposed for free-hand sketches. Strokes can be thickened or dilated~\cite{Xu2020DeepLF} and selectively dropped~\cite{Yu2016SketchaNetAD}, either randomly or based on their positioning on the sketch sequence. Authors in~\cite{Zheng2019SketchSpecificDA} deform training sketches via a Bezier pivot-based deformation strategy (BPD) and \cite{Liu2019AnUS} augment sketches by incorporating random strokes sampled from other sketches. Note that free-hand sketches are distinct from CAD sketches as they follow a stroke-based representation that is non-parametric and limited in terms of editability. In contrast, training augmentations on CAD sketches are mostly overlooked. This work proposes Constraint-Preserving Transformations (CPTs) as a novel strategy for synthetically augmenting CAD sketch datasets.

\vspace{-0.1cm}
\section{Method}
\label{sec:method}

Given a binary sketch image $\mathbf{X} \in \{0,1\} ^{h \times w}$, where $h$ and $w$ denote the height and the width, respectively, our goal is to infer a sketch graph $ \mathcal{G} =(\mathcal{P}^n, \mathcal{C}^m)$ with nodes~\hbox{$\{\mathbf{p}_1, \mathbf{p}_2, ..., \mathbf{p}_n\} \in \mathcal{P}^n$} representing primitives 
 and edges $\{\mathbf{c}_1, \mathbf{c}_2, ..., \mathbf{c}_m\} \in \mathcal{C}^m$ denoting geometric constraints. Similar to \cite{seff2022vitruvion}, $\mathbf{p}_i \in \mathcal{P}$ can be one of the types shown in Table~\ref{tab:types}.  We represent each $\mathbf{p}_i$ by a set of 8 tokens $t^k$, where $t^1\in \llbracket 1..5 \rrbracket$\footnote{$\llbracket \alpha..\beta \rrbracket$ denotes and interval of integers between $\alpha$ and $\beta$.} represents the primitive type (arc, circle, line, point, or no primitive), tokens $t^{ \llbracket 2 .. 7 \rrbracket} \in \llbracket 1..64 \rrbracket$ correspond to quantized primitive parameters and $t^8\in \llbracket 0..1 \rrbracket$ captures whether the primitive is a construction primitive.

\begin{table*}[h]
\setlength{\belowcaptionskip}{-0.3cm}
    \centering
    \setlength{\tabcolsep}{8pt}
    \resizebox{0.8\linewidth}{!}{
    \begin{tabular}{cccc}
        \toprule
        Arc $\mathbf{a}_i$ & Circle $\mathbf{c}_i$ & Line $\mathbf{l}_i$ & Point $\mathbf{d}_i$\\
        $(x_{s}, y_{s},x_{m}, y_{m},x_{e}, y_{e}) \in \mathbb{R}^6$  &
        $(x_{c}, y_{c},r) \in \mathbb{R}^3$  &
        $(x_{s}, y_{s},x_{e}, y_{e}) \in \mathbb{R}^4$ &
        $(x_p, y_p) \in \mathbb{R}^2$ \\ 
        \bottomrule
    \end{tabular}
    }
    \vspace{0.3cm}
    \caption{Primitives types of \texttt{DAVINCI} and corresponding parametarization.}
\label{tab:types}
\end{table*}

 A constraint $\mathbf{c} \in \mathcal{C}$ represents an \textit{undirected} edge between primitives $\mathbf{p}_i$ and $\mathbf{p}_j$. CAD constraint $\mathbf{c}$ also includes subreferences $(s_{i}, s_{j}) \in \llbracket 1..4 \rrbracket^2$, to specify whether the constraint is applied on start, end, middle point, or entire primitive for both $\mathbf{p}_i$ and $\mathbf{p}_j$. Each constraint is characterized by a relation label $c_{i,j}^{s_{i},s_{j}} \in \llbracket 1..T \rrbracket$, which denotes the type of the constraint (\eg coincident, parallel, midpoint, etc.), where $T$ is the total number of considered constraint types. Note that some constraints may involve only a single primitive $\mathbf{p}_i$; in such cases, the constraint is defined as the edge between the primitive and itself. For further tokenization details readers are referred to~\cite{seff2022vitruvion}.

\subsection{CAD Sketch Graph Inference} 
\begin{figure*}[t]
    \setlength{\belowcaptionskip}{-0.4cm}
    \centering
    \includegraphics[width=0.98\textwidth]{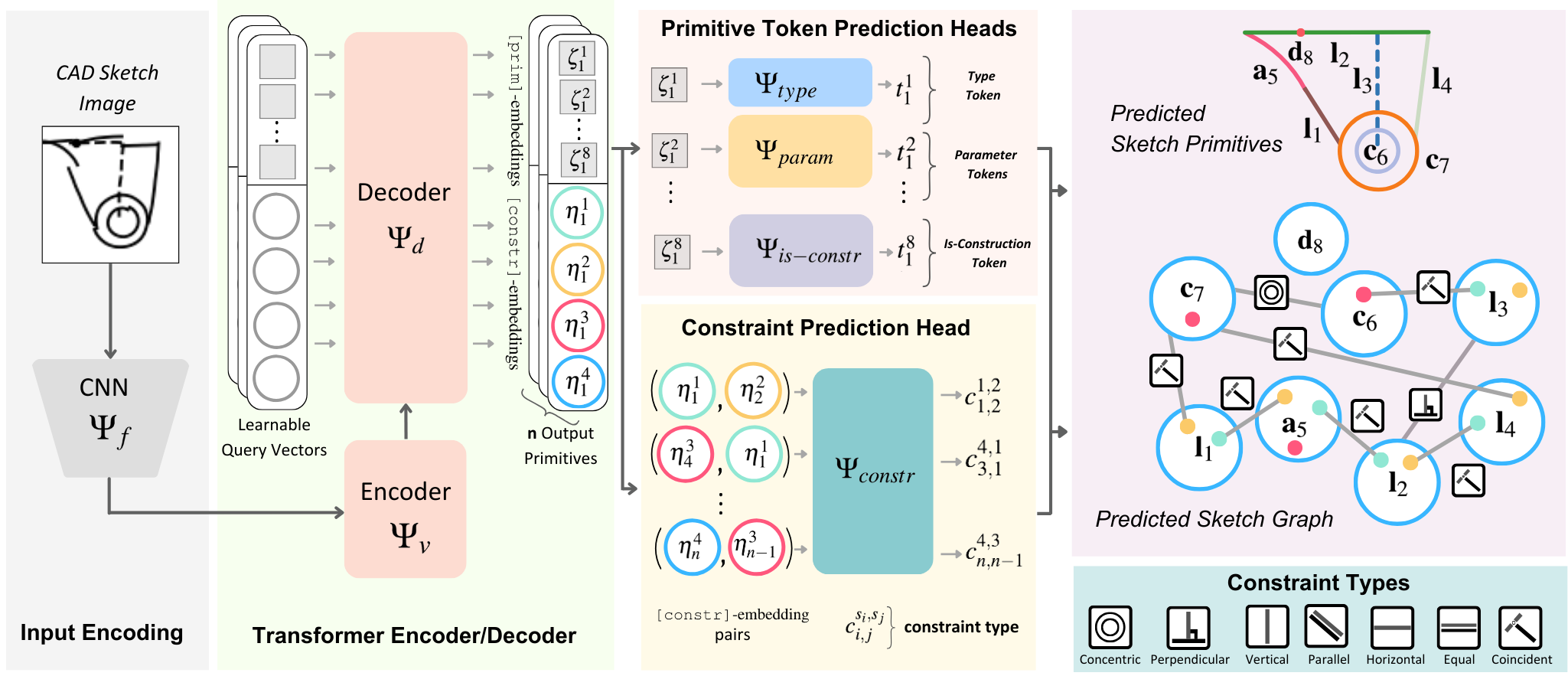}
    \vspace{0.3cm}
    \caption{Overview of the \texttt{DAVINCI} architecture. A raster CAD sketch image is processed by a transformer encoder/decoder. Predicted \texttt{[prim]}-embeddings and \texttt{[constr]}-embeddings are used to infer primitive tokens and geometric constraints between primitives.}
    \label{fig:spn}
\end{figure*}

\texttt{DAVINCI}  is an encoder/decoder transformer network that jointly learns primitive parameterizations and their constraint relationships (\textit{i.e.} sketch graph $(\mathcal{P}^n, \mathcal{C}^m)$) from a raster CAD sketch image $\mathbf{X}$. To that end, it learns two types of token embeddings; \texttt{[prim]}-embeddings and \texttt{[constr]}-embeddings. A \texttt{[prim]}-embedding is learned for each primitive token and can be directly decoded to the token value $t^k$ via dedicated MLP heads. \texttt{[constr]}-embeddings are utilized for constraint inference. For each primitive, we learn $4$ \texttt{[constr]}-embeddings, each corresponding to a different primitive subreference. A constraint prediction MLP head is used on the pair-wise combinations of these \texttt{[constr]}-embeddings to infer the existence of geometric constraints between subreferences of primitives. Note that the \texttt{[constr]}-embeddings are learned per primitive (\textit{i.e.} node) and not per constraint (\textit{i.e.} undirected edge). Such an alternative would require a total of $4n(4n-1)/2$ \texttt{[constr]}-embeddings (\textit{i.e.} quadratic \textit{w.r.t} to the number of nodes), whereas the proposed approach only requires $4n$ \texttt{[constr]}-embeddings (\textit{i.e.} linear \textit{w.r.t} to the number of nodes). This modeling approach is motivated by the recently proposed Relationformer~\cite{shit2022relationformer} for image-to-graph generation. Compared to~\cite{shit2022relationformer}, \texttt{DAVINCI} infers multiple relationship embeddings to query different types of possible primitive subreferences. An overview of \texttt{DAVINCI} is depicted in Figure~\ref{fig:spn}.

\subsection{\texttt{DAVINCI} Architecture}  First, a convolutional backbone $\Psi_f$ extracts a set of image features $\mathbf{f} \in \mathbb{R} ^{d_f \times h \times w}$
for input sketch image $\mathbf{X}$. The input feature map $\mathbf{f}$ is combined with fixed positional embeddings~\cite{Parmar2018ImageT} and processed by a transformer encoder $\Psi_v$. Then, a decoder $\Psi_d$ maps learnable query vectors into $8n$ \texttt{[prim]}-embeddings $\{\mathbf{\zeta}_i^k \}^{k \in \llbracket 1..8 \rrbracket}_{i \in \llbracket 1..n \rrbracket} \in \mathbb{R}^{d_e}$ and $4n$ \texttt{[constr]}-embeddings $\{\mathbf{\eta}_i^k \}^{k \in \llbracket 1..4 \rrbracket}_{i \in \llbracket 1..n \rrbracket} \in \mathbb{R}^{d_e}$ via cross attention on the features encoded by $\Psi_v$. Learned embeddings are processed by two categories of output heads, the \textit{primitive token prediction heads} and \textit{constraint prediction head.}

\vspace{0.1cm}
\noindent
\textbf{Primitive Token Prediction Heads.} Each \texttt{[prim]}-embedding $\mathbf{\zeta}_i^k$ is mapped to the corresponding primitive token values via 3-layer MLPs. We use 3 heads $\Psi_{type}: \mathbb{R}^{d_e} \rightarrow \llbracket 1..5 \rrbracket$ for type token prediction, $\Psi_{param}: \mathbb{R}^{d_e} \rightarrow \llbracket 1..64 \rrbracket$ for token parameters, and $\Psi_{is-constr}: \mathbb{R}^{d_e} \rightarrow \llbracket 0..1 \rrbracket$ for the is-construction flag.

\vspace{0.1cm}
\noindent
\textbf{Constraint Prediction Head.} This head consisting of a 3-layer MLP, $\Psi_{constr}$, which infers the existence of a geometric constraint within a pair of \texttt{[constr]}-embeddings from different primitives. Since constraint relationships are undirected, we ensure permutation invariance on the input pair of \texttt{[constr]}-embeddings. This is done by concatenating the sum and subtraction of the two embeddings, \textit{i.e.} $[\mathbf{\eta}_i^k + \mathbf{\eta}_j^k, \mathbf{\eta}_i^k - \mathbf{\eta}_j^k]$. We then learn the mapping $\Psi_{constr}: \mathbb{R}^{2d_e} \rightarrow \llbracket 1..(T+1) \rrbracket$ where $T+1$ are the number of constraints plus the $\varnothing$-constraint class.

\subsection{\texttt{DAVINCI} Training Loss} \texttt{DAVINCI} is trained for joint primitive parameterization and constraint inference. Primitive tokens are learned via a standard cross-entropy classification loss between predicted $\hat{t}_i^k$ and ground truth $t_i^k$ tokens. Note that as a set-based transformer model, one-to-one correspondence $\hat{\pi} \in \Pi_n$ between predicted and ground truth primitive tokens, is computed through optimal bipartite matching via the Hungarian matching~\cite{kuhn1955hungarian} algorithm (as in \cite{carion2020end}). Here, $\Pi_n$ denotes the space of all bijections from the set $\llbracket 1..n \rrbracket$ to itself. To predict geometric constraints, we also employ a cross-entropy loss between predicted $\hat{c}_{i,j}^{s_i,s_j}$ and ground truth constraints $c_{i,j}^{s_i,s_j}$. Note that the obtained primitive bipartite matching is used on constraint loss computation. Given that geometric constraints are highly sparse, most pairwise \texttt{[constr]}-embedding combinations would be categorized as $\varnothing$-constraints. To ensure that the learning space will be limited to only valid subreference combinations, we select \texttt{[constr]}-embedding pairs conditioned on the corresponding primitive types. In particular,  we define the valid combinations \mbox{$S = \left\{ (i, s_i, j, s_j) \mid i, j  \in \llbracket 1..n \rrbracket, \, s_i \in \mathcal{V}(t_i^1), \, s_j \in \mathcal{V}(t_j^1) \right\}$}, where  $ \mathcal{V}(1) \in \llbracket 1..4 \rrbracket $, $\mathcal{V}(2) \in \{2,3\}$, $\mathcal{V}(3) \in \{1,2,4\}$ and  $\mathcal{V}(4) \in \{4\}$ for primitive types $t_i^1$ (\textit{i.e.} arc, circle, line, and point). Following this strategy, topologically invalid pairs (\eg $<$\textit{circle}.\texttt{start} - \textit{line}.\texttt{end} $>$) would be removed from the loss computation. The total loss has the form,

\begin{equation}
    \mathcal{L}_{total} = \sum_{i=1}^n \sum_{k=1}^8 \mathcal{L}_{cls}(t_i^k, \hat{t}_{\hat{\pi}(i)}^k) \hspace{0.2cm} + \
    \sum_{\{i, s_i, j, s_j\} \in ~\mathcal{S}} \mathcal{L}_{cls}(c_{i,j}^{s_i,s_j},\hat{c}_{\hat{\pi}(i),\hat{\pi}(j)}^{s_i,s_j}) \ , 
    \label{loss}
\end{equation}

\noindent where $\mathcal{L}_{cls}$ is the standard cross-entropy classification loss. During constraint inference, we similarly use the valid combination set $\mathcal{S}$ (conditioned on the predicted primitive type $\hat{t}^1_i$).

\vspace{-0.2cm}
\section{Constraint Preserving Transformations}
\label{sec:cpt}

\begin{figure}[t]
\setlength{\belowcaptionskip}{-0.4cm}
    \centering
\includegraphics[width=0.9\linewidth]{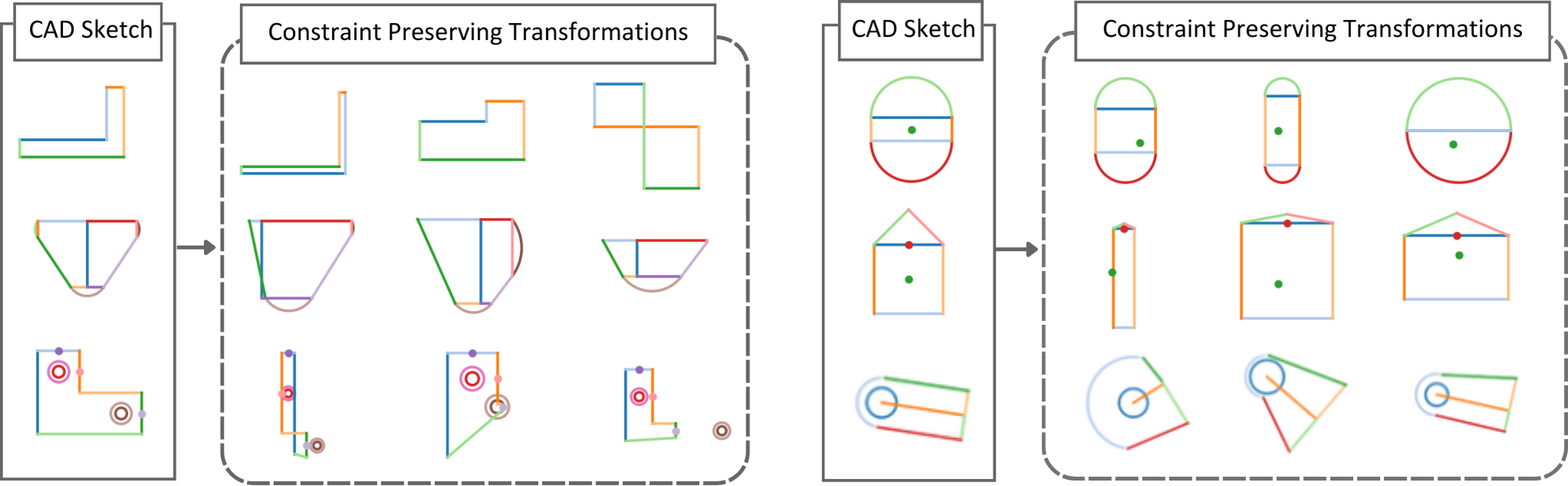}
    \vspace{0.3cm}
    \caption{Example Constraint-Preserving Transformations (CPTs) of CAD sketches from \textit{CPTSketchGraphs}. CPTs are generated via intergration with the FreeCAD API~\cite{FreeCAD}. }
    \label{fig:cpt_examples}
\end{figure}

Constraint-Preserving Transformations (CPTs) introduce a novel approach for data augmentation in CAD sketches. The core idea is to leverage the existing geometric constraints within sketches to generate plausible variations of the original design. The augmentation process is enabled via integration with FreeCAD software's API~\cite{FreeCAD}. A random local perturbation, such as the translation of a sketch point, is automatically applied to the CAD sketch. Due to the constraints associated with the manipulated point, this local change cascades across the sketch, modifying the parameterization of all connected primitives. Importantly, these transformations preserve the original geometric and topological relationships set by the constraints. The rationale behind CPTs stems from the foundational concept of design intent in CAD, which is intrinsically linked to how a sketch reacts to modifications~\cite{Otey2018RevisitingTD}. During the design process, CAD engineers strategically establish constraints to enable specific degrees of freedom while suppressing others, thus guiding the permissible parametrizations of the sketch. In essence, CPTs synthesize potential CAD sketches that remain within the boundaries defined by the designer, ensuring that the generated sketches are valid instances under the original design specifications. 

\begin{wrapfigure}{r}{0.65\textwidth}
    \setlength{\belowcaptionskip}{-0.4cm}
  \centering
  \includegraphics[width=0.6\textwidth]{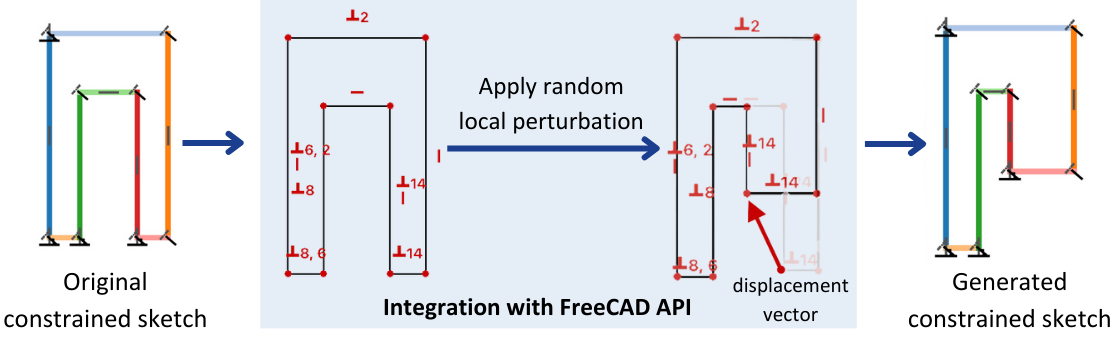}
  \vspace{0.3cm}
  \caption{Formation of a CPT via random local perturbations. }
  \label{fig:cpt_generation}
\end{wrapfigure}

To generate CPTs, we perform $M$ subsequent perturbations on randomly sampled sketch subreferences (see Figure~\ref{fig:cpt_generation}). Each perturbation involves a displacement vector that shifts the original subreference to a new location within a predefined bounding box, denoted as $bbox_{disp}$. This bounding box is configured as a square, centered at the subreference's original coordinate. The width $w_b$ of $bbox_{disp}$, is dependent to the extent of the sampled primitive. It is important to note that not all displacements will necessarily result in novel sketch parameterizations, since some CAD sketches may be fully constrained (\textit{i.e.} having no degrees of freedom).  Some examples of CPTs are illustrated in Figure~\ref{fig:cpt_examples}. By applying CPTs on SketchGraphs dataset~\cite{seff2020sketchgraphs}, originally formed by 1.5 million   CAD sketches, we introduce \textit{CPTSketchGraphs}, a dataset comprising 80 million CAD sketches.

\vspace{-0.3cm}
\section{Experiments}
\label{sec:experiments}

\vspace{-0.3cm}
\noindent
\textbf{Dataset.} We evaluate the performance of \texttt{DAVINCI} on the SketchGraphs dataset~\cite{seff2020sketchgraphs}. We follow the preprocessing of~\cite{seff2022vitruvion} to remove duplicates and simplistic sketches. The dataset includes $1.53$ million samples. We use the train/val splits of~\cite{seff2022vitruvion} and perform testing on a set of $5000$ sketches. Hand-drawn sketches are generated via the synthesis process of~\cite{seff2022vitruvion}.  \texttt{DAVINCI} considers the following types of constraints: \textit{coincident, concentric, equal, fix, horizontal, midpoint, normal, offset, parallel, perpendicular, quadrant, tangent, vertical}.

\vspace{0.1cm}
\noindent
\textbf{Implementation Details.} Sketch images are of dimension $128\times128$. For the backbone $\Psi_f$, we employ a U-Net~\cite{Ronneberger2015UNetCN} with a ResNet34~\cite{He2015DeepRL} encoder that produces a $16$ dimensional feature map. The transformer modules $\Psi_v$ and $\Psi_d$ are formed by $4$ layers each, with $8$ heads and $256$ latent dimensions. As in~\cite{seff2022vitruvion}, the maximum number of primitives is fixed to $n=16$. \texttt{DAVINCI} is trained for $90$ epochs with a batch size of $2048$ and a learning rate of $3.5\times10^{-4}$, that is scaled by a factor of $0.3$ after $80$ epochs. We implemented all modules in Pytorch~\cite{Paszke2017AutomaticDI}. An epoch takes approximately $30$ mins on 4 NVIDIA A100's.

\vspace{0.1cm}
\noindent
\textbf{Evaluation.} 
For evaluating primitive recovery, we utilize three metrics; Accuracy \textit{(Acc)}, Primitive F1 Score \textit{(PF1)} and bidirectional Chamfer Distance \textit{(CD)}. Accuracy is calculated \textit{w.r.t} the ground truth token sequence. \textit{PF1} considers a true-positive to be a primitive with correctly predicted type and all parameters within $5$ quantization units as in~\cite{yang2022discovering}. For \textit{CD} computation, we sample points uniformly on predicted and ground truth primitives. Constraint prediction is evaluated in terms of Constraint F1 Score \textit{(CF1)} as in~\cite{yang2022discovering}. A constraint relationship is considered a true-positive only if all involved primitives are also true-positives. Note that since \texttt{DAVINCI} is a set-based method, correspondence \textit{w.r.t} the ground truth is recovered prior to evaluation.

\begin{figure}
\setlength{\belowcaptionskip}{-0.5cm}
    \centering
\includegraphics[width=0.95\linewidth]{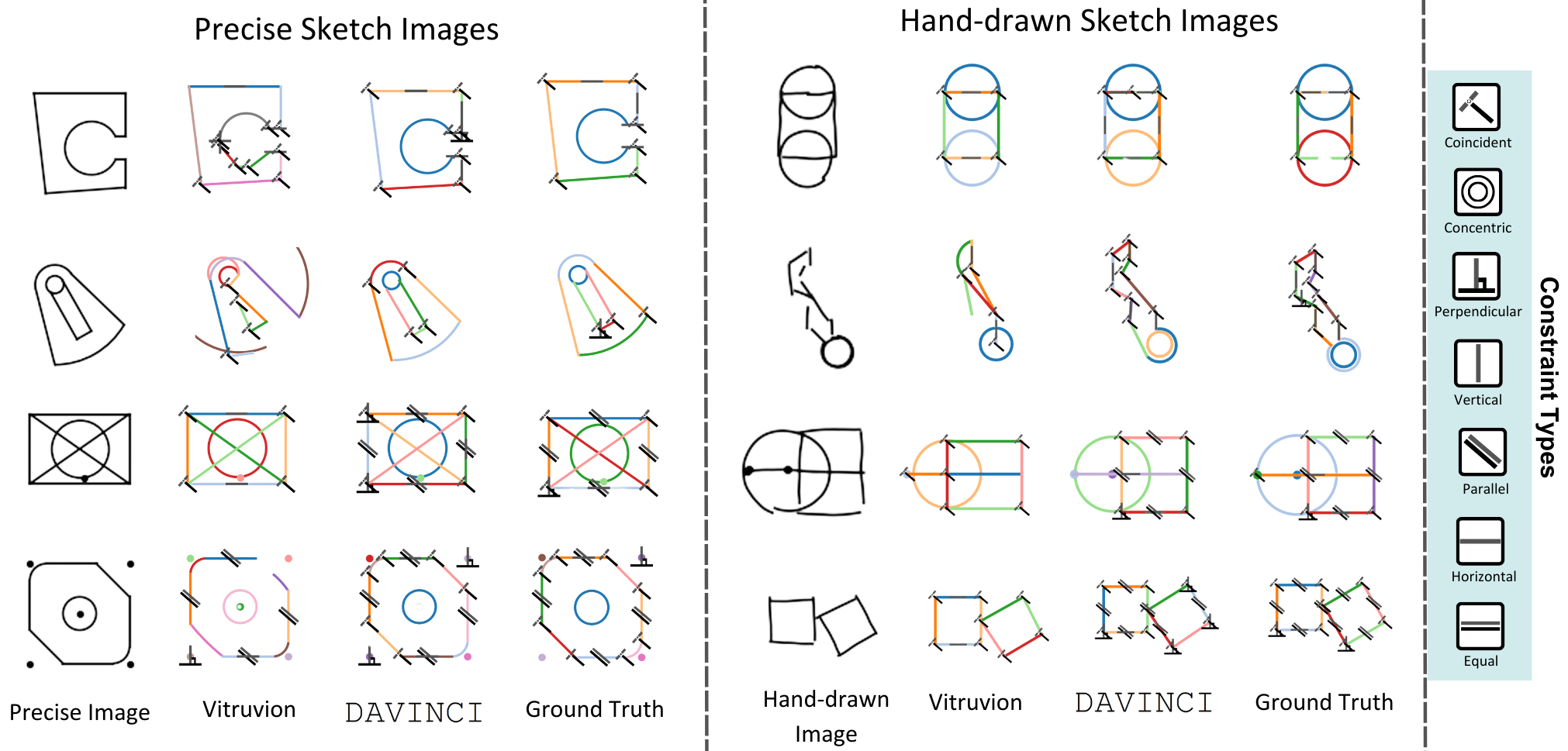}
    \vspace{0.2cm}
    \caption{Qualitative comparison with Vitruvion~\cite{seff2022vitruvion}.}
    \label{fig:results_qualitative}
    \vspace{0.2cm}
\end{figure}

\begin{table*}[t]
\setlength{\belowcaptionskip}{-0.4cm}
    \centering
    \setlength{\tabcolsep}{4pt}
    \resizebox{0.8\linewidth}{!}{
    \begin{tabular}{lcccccccc}
        \toprule
        \multicolumn{1}{c}{} & \multicolumn{4}{c}{\textbf{Precise Sketch Images}} & \multicolumn{4}{c}{\textbf{Hand-drawn Sketch Images}} \\
        \cmidrule(r){2-5} \cmidrule(l){6-9}
        Method & \textit{Acc} $\uparrow$ & \textit{CD} $\downarrow$ & \textit{PF1} $\uparrow$ & \textit{CF1} $\uparrow$ & \textit{Acc} $\uparrow$ & \textit{CD} $\downarrow$ & \textit{PF1} $\uparrow$ & \textit{{CF1}} $\uparrow$  \\
        \midrule
        Vitruvion~\cite{seff2022vitruvion} & 0.8228 & 0.2702 & 0.8302 & 0.1915 & 0.6736 & 0.9271 & 0.6976 & 0.2392 \\
        \texttt{DAVINCI} & \textbf{0.8826} & \textbf{0.2650} & \textbf{0.9172} & \textbf{0.6281} & \textbf{0.8315} & \textbf{0.5966} & \textbf{0.8801} & \textbf{0.6284}  \\
        \bottomrule
    \end{tabular}
    }
    \vspace{0.3cm}
    \caption{Quantitative comparison with Vitruvion on the SketchGraphs dataset~\cite{seff2020sketchgraphs}. Both methods are trained on precise and hand-drawn sketches.}
\label{tab:results_vitruvion}
\end{table*}

\vspace{-0.3cm}
\subsection{Constrained CAD Sketch Inference}
We start by evaluating the effectiveness of \texttt{DAVINCI} for joint primitive and constraint inference from CAD sketch raster images. Comparison in Table~\ref{tab:results_vitruvion} is performed with the state-of-the-art method of Vitruvion~\cite{seff2022vitruvion} on both precise and hand-drawn images. Both \texttt{DAVINCI} and Vitruvion~\cite{seff2022vitruvion} are trained on SketchGraphs. To assess the proposed \texttt{DAVINCI} architecture, we do not include CPT augmentations for this experiment.  

\begin{wraptable}{r}{0.4\linewidth} %
    \centering
    \setlength{\belowcaptionskip}{-0.3cm}
    \setlength{\tabcolsep}{4pt}
    \resizebox{\linewidth}{!}{
    \begin{tabular}{lcc}
        \toprule
        Method & \textit{PF1} $\uparrow$ & \textit{CF1} $\uparrow$  \\
        \midrule
        SketchConcepts~\cite{yang2022discovering} & 0.711 &  0.368 \\
        \texttt{DAVINCI} & \textbf{0.822} & \textbf{0.601} \\
        \bottomrule
    \end{tabular}
    }
    \vspace{0.2cm}
    \caption{Comparison with~\cite{yang2022discovering} on hand-drawn CAD sketches.}
    \label{tab:results_sketchconcepts}
\end{wraptable}

The proposed \texttt{DAVINCI} outperforms Vitruvion~\cite{seff2022vitruvion} at both primitive and constraint levels on all metrics by a large margin. A qualitative comparison of the two methods is provided in Figure~\ref{fig:results_qualitative}. We additionally compare \texttt{DAVINCI} to SketchConcepts~\cite{yang2022discovering} on a hand-drawn setting in
Table~\ref{tab:results_sketchconcepts}. For this experiment, we utilize the same splits as~\cite{yang2022discovering}, which differ from the splits used in Table~\ref{tab:results_vitruvion}. Note that~\cite{yang2022discovering} detects constraints as relationships between primitives without predicting constraint subreferences (\eg start, mid, end points). Therefore, we adapted the calculation of the \textit{CF1} to exclude subreferences in the true-positive computation. Our proposed method surpasses SketchConcepts\footnote{Results for SketchConcepts are taken directly from the supplementary material of the paper~\cite{yang2022discovering}, as the implementation of constrained CAD sketch parameterization from images is not publicly available.}~\cite{yang2022discovering} by a significant margin.

\vspace{-0.2cm}
\subsection{Effectiveness of the Constraint Preserving Transformations}

\begin{table*}[t]
\setlength{\belowcaptionskip}{-0.7cm}
    \centering
    \setlength{\tabcolsep}{4pt}
    \resizebox{\linewidth}{!}{
        \begin{tabular}{llcccccccc}
            \toprule
            \multicolumn{1}{c}{} &\multicolumn{1}{c}{} & \multicolumn{4}{c}{\textbf{Precise Sketch Images}} & \multicolumn{4}{c}{\textbf{Hand-drawn Sketch Images}} \\
            \cmidrule(r){3-6} \cmidrule(l){7-10}
            Method & Dataset Size &\textit{Acc} $\uparrow$ & \textit{CD} $\downarrow$ & \textit{PF1} $\uparrow$ & \textit{CF1} $\uparrow$ & \textit{Acc} $\uparrow$ & \textit{CD} $\downarrow$ & \textit{PF1} $\uparrow$ & \textit{CF1} $\uparrow$ \\
            \midrule
            upper bound / no augmentations & 100k & 0.7792 & 1.3294 & 0.7082 & 0.3725 & 0.7065 & 1.4811 & 0.6824 & 0.3746\\
            \midrule
            lower bound / no augmentations & 1k & 0.4845 & 5.5323 & 0.1081 & 0.0108  & 0.4808 & 4.8397 & 0.1250 & 0.0120 \\
            \textit{synthetic CAD sketches} & 1k & 0.5989 & 2.9880 & 0.3536 & 0.1152 & 0.5241 & 2.9955 & 0.3650 & 0.1393 \\
            \textit{rotated CAD sketches} & 1k & 0.6603 & 2.4751 & 0.4607 & 0.1756 & 0.5656 & 2.7377 & 0.4079 & 0.1261 \\
            CPTs& 1k& \textbf{0.7006} & \textbf{1.9148} &  \textbf{0.5791} & \textbf{0.2882} & \textbf{0.6182} & \textbf{2.0164} & \textbf{0.5508} & \textbf{0.2760} \\
            \bottomrule
        \end{tabular}
        }
    \vspace{0.2cm}
    \caption{Comparison of different CAD sketch augmentation strategies.}
\label{tab:results_augmentations}
\end{table*}

This section evaluates the effectiveness of the proposed Constraint Preserving Transformations (CPTs) as an augmentation strategy for constrained CAD sketch parameterization. We compare CPTs with two augmentation strategies; \textit{synthetic} and \textit{rotated} CAD sketches.

\begin{wrapfigure}{l}{0.5\textwidth}
\setlength{\belowcaptionskip}{-0.3cm}
  \centering
  \includegraphics[width=0.5\textwidth]{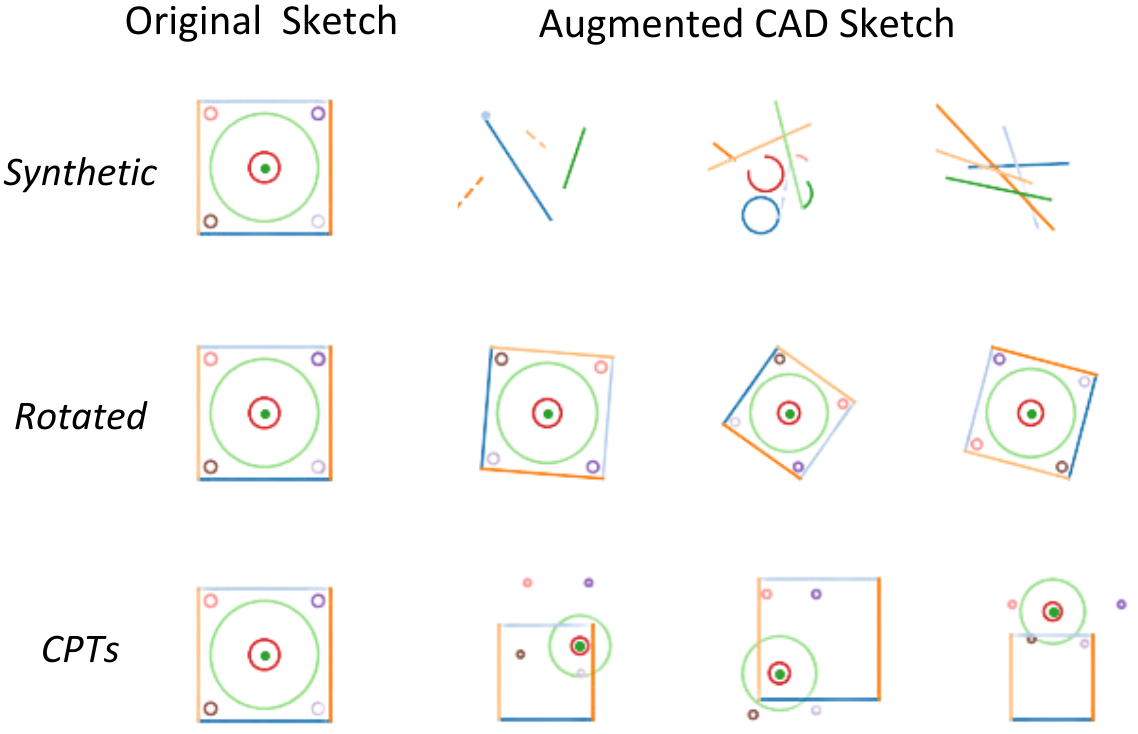}
  \vspace{0.3cm}
  \caption{Investigated augmentation strategies.}
  \label{fig:sketch_augmentations}
\end{wrapfigure}

Synthetic sketches are built iteratively by appending sampled primitives and constraints from a predefined set of sketch patterns (\eg adding a new random line that remains parallel to an existing line or a new circle that is concentric to an existing circle). For rotated CAD sketches, we exclude the constraints that become invalid (\eg horizontal) due to the sketch rotation. Some visual examples of the investigated augmentation strategies are depicted in Figure~\ref{fig:sketch_augmentations}. To investigate the effectiveness of different augmentation strategies, we conduct an experiment where \texttt{DAVINCI} is trained on a small set of $1000$ CAD sketches from SketchGraphs~\cite{seff2020sketchgraphs} ($\approx 0.1\%$ of the full dataset). We generate $100$ augmentations per sketch for each augmentation method, leading to a training set of $100k$ samples. During training, the model is trained with both augmented sketches (sampled with a probability of $0.7$) and original samples. As an upper bound, we also report the performance of a model trained with $100k$ samples from SketchGraphs. Results are summarized in  Table~\ref{tab:results_augmentations}. We observe that when trained without any augmentations (lower bound), \texttt{DAVINCI} completely underperforms. Both augmentation baseline methods enhance model performance at primitive and constraint levels. Overall, CPT-based augmentations achieve the best results, reaching a comparable performance \textit{w.r.t} the upper bound ($\approx 89\%$ of upper bound \textit{Acc} and $\approx75\%$ of \textit{CF1}).

\vspace{-0.1cm}
\subsection{Ablation Study and Future Reverse Engineering Applications}

\noindent \textbf{Ablation Study.} We conduct an ablation study on the effect of joint primitive parameterization and constraint inference that is enabled by \texttt{DAVINCI}. Existing methods treat primitive and constraint inference as separate problems addressed via two-stage pipelines~\cite{para2021sketchgen, seff2022vitruvion}. This decomposition can lead to diminished performance, as errors in primitive prediction adversely affect subsequent constraint inference. In Table~\ref{tab:ablation_constraints}, we report the performance of the constraint prediction model of Vitruvion~\cite{seff2022vitruvion}, for different input primitive parameterizations. We observe that the constraint prediction performance of \cite{seff2022vitruvion} is improved when initial primitive prediction is provided by \texttt{DAVINCI}, particularly on precise sketches. Additionally, jointly training primitives and constraints enabled by \texttt{DAVINCI}, helps prevent the accumulation of errors and achieves superior constraint inference performance across both precise and hand-drawn images. The effect of joint training is less apparent on primitive parameterization performance. We observe in Table~\ref{tab:ablation_parameterization}, that a variant of \texttt{DAVINCI} trained with the primitive parameterization loss only, can be on-par with the joint training model.

\begin{table*}[t]
\setlength{\belowcaptionskip}{-0.3cm}
    \centering
    \setlength{\tabcolsep}{4pt}
    \resizebox{0.8\linewidth}{!}{
    \begin{tabular}{c|ccc}
        \toprule
        Primitive Prediction & Constraint Prediction &  \textit{Precise CF1} $\uparrow$ & \textit{Hand-drawn CF1} $\uparrow$ \\
        \midrule
        Vitruvion~\cite{seff2022vitruvion} & Vitruvion~\cite{seff2022vitruvion} &   0.1915  & 0.2392\\
        \texttt{DAVINCI} & Vitruvion~\cite{seff2022vitruvion}  & 0.2677  & 0.2349 \\
        \multicolumn{2}{c}{\texttt{DAVINCI}\textit{(joint inference)}} & \textbf{0.6281}  & \textbf{0.6284} \\
        \bottomrule
    \end{tabular}
    }
    \vspace{0.3cm}
    \caption{Ablation study on constraint inference.}
\label{tab:ablation_constraints}
\end{table*}

\begin{table*}[t]
\setlength{\belowcaptionskip}{-0.5cm}
    \centering
    \setlength{\tabcolsep}{4pt}
    \resizebox{0.8\linewidth}{!}{
    \begin{tabular}{lcccccc}
        \toprule
        \multicolumn{1}{c}{} & \multicolumn{3}{c}{\textbf{Precise Sketch Images}} & \multicolumn{3}{c}{\textbf{Hand-drawn Sketch Images}} \\
        \cmidrule(r){2-4} \cmidrule(l){5-7}
        Method & \textit{Acc} $\uparrow$ & \textit{CD} $\downarrow$ & \textit{PF1} $\uparrow$ & \textit{Acc} $\uparrow$ & \textit{CD} $\downarrow$ & \textit{PF1} $\uparrow$\\
        \midrule
        \texttt{DAVINCI} w/o Constraints & 0.8791 & 0.3147 & 0.9110  & 0.8230 & 0.5946 & 0.8729 \\
        \texttt{DAVINCI} & \textbf{0.8826} & \textbf{0.2650} &  \textbf{0.9172} & \textbf{0.8315} &  \textbf{0.5966} & \textbf{0.8801}\\
        \bottomrule
    \end{tabular}
    }
    \vspace{0.3cm}
    \caption{Ablation study on primitive parameterization.}
\label{tab:ablation_parameterization}
\end{table*}

\begin{figure}[h]
\setlength{\belowcaptionskip}{-0.7cm}
    \centering
\includegraphics[width=\linewidth]{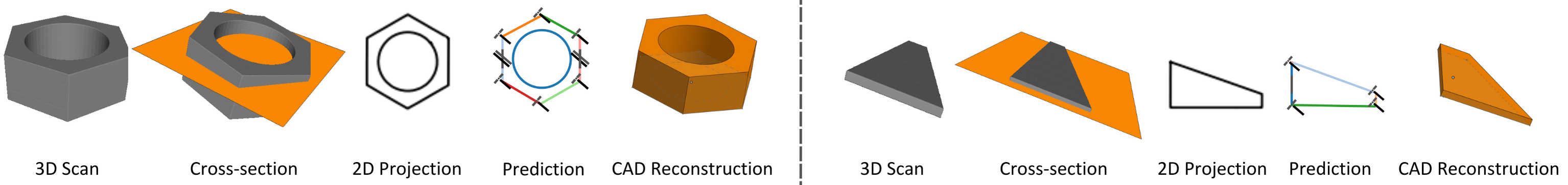}
    \vspace{0.01cm}
    \caption{Parameterization of 2D cross-sections.}
    \label{fig:results_cross_section}
\end{figure}

\vspace{0.1cm}
\noindent \textbf{Future Applications.} In addition to precise and hand-drawn inputs, \texttt{DAVINCI} can also enable constrained CAD sketch inference on 2D cross-sections of 3D scans. Cross-sections are critical components of CAD reverse engineering pipelines~\cite{cross_sections} and are obtained by designers via intersecting an existing 3D scan with a selected 2D plane. Figure~\ref{fig:results_cross_section} presents preliminary qualitative examples of \texttt{DAVINCI} applicability within this reverse engineering scenario. Note that on this setting, \texttt{DAVINCI} is trained on SketchGraphs. The quantitative evaluation is not included as there is no annotated cross-section dataset available.

\vspace{-0.1cm}
\section{Conclusion and Future Works}
\label{sec:conclusion}

This paper presents \texttt{DAVINCI}, a novel single-stage architecture for constrained CAD sketch parameterization from CAD sketch raster images. The proposed method achieves the state-of-the-art performance on both precise and hand-drawn CAD sketch images. We also propose Constraint Preserving Transformations (CPTs) as a novel augmentation strategy tailored to CAD sketches that allows effective model performance under limited data scenarios. We believe that the proposed CPTs can be beneficial for large-scale training of constrained CAD sketch parameterization models. Such investigation, enabled by the proposed \textit{CPTSketchGraphs} dataset of $80$ million CPTs, is left for future work. Building a dataset of cross-sections and validating \texttt{DAVINCI} on it is also left for future investigations.

\section{Acknowledgements}
The present work is supported by the National Research Fund (FNR), Luxembourg, under the BRIDGES2021/IS/16849599/FREE-3D project and by Artec3D.

\bibliography{main}
\end{document}